\documentclass[10pt,twocolumn,letterpaper]{article}

\usepackage{cvpr}              

\usepackage{graphicx}
\usepackage{amsmath}
\usepackage{amssymb}
\usepackage{mathtools}
\usepackage{amsthm}
\usepackage{multirow}
\usepackage{balance}
\usepackage{bm}
\usepackage{amssymb}
\usepackage{adjustbox}
\usepackage{booktabs}
\usepackage{caption}
\usepackage{blindtext}   
\usepackage{algorithm}
\usepackage{algorithmicx, algpseudocode}
\usepackage[table,xcdraw]{xcolor}
\usepackage{pifont}

\usepackage[normalem]{ulem}
\useunder{\uline}{\ul}{}

\definecolor{cvprblue}{rgb}{0.21,0.49,0.74}
\usepackage[pagebackref,breaklinks,colorlinks,citecolor=cvprblue]{hyperref}

\def\paperID{9} 
\def\confName{CVPR}
\def\confYear{2024}

\title{HD Maps are Lane Detection Generalizers: \\A Novel Generative Framework for Single-Source Domain Generalization
}

\author{Daeun Lee\thanks{Work done during the research internship at NAVER LABS.}\\
Korea University\\
{\tt\small goodgpt@korea.ac.kr}
\and
Minhyeok Heo\\
NAVER LABS\\
{\tt\small heo.minhyeok@naverlabs.com}
\and
Jiwon Kim\\
NAVER LABS\\
{\tt\small g1.kim@naverlabs.com}
}

\begin{document}
\maketitle

\begin{abstract}
Lane detection is a vital task for vehicles to navigate and localize their position on the road. 
To ensure reliable driving, lane detection models must have robust generalization performance in various road environments. 
However, despite the advanced performance in the trained domain, their generalization performance still falls short of expectations due to the domain discrepancy.
To bridge this gap, we propose a novel generative framework using HD Maps for Single-Source Domain Generalization (SSDG) in lane detection. 
We first generate numerous front-view images from lane markings of HD Maps. 
Next, we strategically select a core subset among the generated images using (i) lane structure and (ii) road surrounding criteria to maximize their diversity. 
In the end, utilizing this core set, we train lane detection models to boost their generalization performance.
We validate that our generative framework from HD Maps outperforms the Domain Adaptation model MLDA with +3.01\%p accuracy improvement, even though we do not access the target domain images. 


\end{abstract}

\section{Introduction}

Lane detection is a fundamental task in autonomous driving. 
By detecting lane markings on the road, vehicles can localize their position and identify drivable spaces. 
However, current data-driven lane detection models still suffer from performance degradation when tested on unseen domains.
Such discrepancies between the source and target datasets result in a severe performance decrease. 


Recent studies~\cite{dg1,dg2,dg3,dg4,dg5} have attempted to address this performance degradation.
These approaches include Domain Adaptation (DA), which involves making use of a limited number of data samples from the target domain, and Domain Generalization (DG), which leverages data from multiple source domains.
However, in realistic driving scenarios, it is technically impossible to obtain even a small number of data samples from the target domain since they cannot be predetermined. 
Collecting and annotating data manually from numerous regions also have practical limitations.
Therefore, it is more realistic to achieve generalization solely relying on a single-source domain without any prior information about the target domain.

From the Single-Source Domain Generalization (SSDG) perspective, it is imperative to obtain domain-invariant features, which necessitates a sufficiently \textit{diverse} set of source domain data. 
To achieve this diversity, previous works~\cite{hendrycks2019augmix, volpi2018generalizing, qiao2020learning, ssdg1, ssdg2, ssdg3} actively adopt augmentation techniques such as Augmix~\cite{hendrycks2019augmix} or adversarial methods~\cite{volpi2018generalizing, qiao2020learning}. 
However, SSDG for lane detection has been unexplored, even though the significance for the real-world application.

Therefore, we propose a \textit{model-agnostic} generative framework to generalize lane detection models adopting \textit{HD Maps} as a \textit{single-source domain}.
To the best of our knowledge, this is the first paper addressing the domain shifts in lane detection with SSDG setup. 
Notably, we notify that HD Maps are effective lane-detection generalizers thanks to their flexibility in collecting numerous datasets depending on the camera parameters.   
Utilizing this advantage, we first get the 2D lane masks from pre-built HD Maps and generate lane-conditioned images with various surroundings. 
Next, we strategically select a core set among the generated images to maximize diversity by using \textit{(i) lane structure} and \textit{(ii) road surrounding} criteria. 
This can help to boost the robustness of our single-source domain from HD Maps generation. 
In the end, we train lane detection models regarding this core set as a single-source domain. 

The main contributions of the paper are summarized as follows:

\begin{itemize}

\item We propose a model-agnostic generative framework to generalize a lane detection model utilizing a single-source domain from HD Maps, addressing the SSDG problem in lane detection for the first time.

\item After decomposing the factor of domain shifts with \textit{(i) lane structure} and \textit{(ii) road surrounding}, we propose the core set selection criteria to maximize the diversity of single-source domain. 

\item Diverse experimental validations have demonstrated that the proposed training framework outperforms previous DA methods without any target domain information.

\end{itemize}



\section{Methodology} 


We first introduce the coreset selection criteria for lane and surroundings in~\Cref{sec:subsec:core}. 
Next, we describe our total framework including the lane-conditioned generation from HD Maps and lane detection model training process in~\Cref{sec:subsec:train}. 

\begin{figure*}[!t]
    \vspace{0.2cm}
    \centering
    \includegraphics[width=0.83\textwidth]{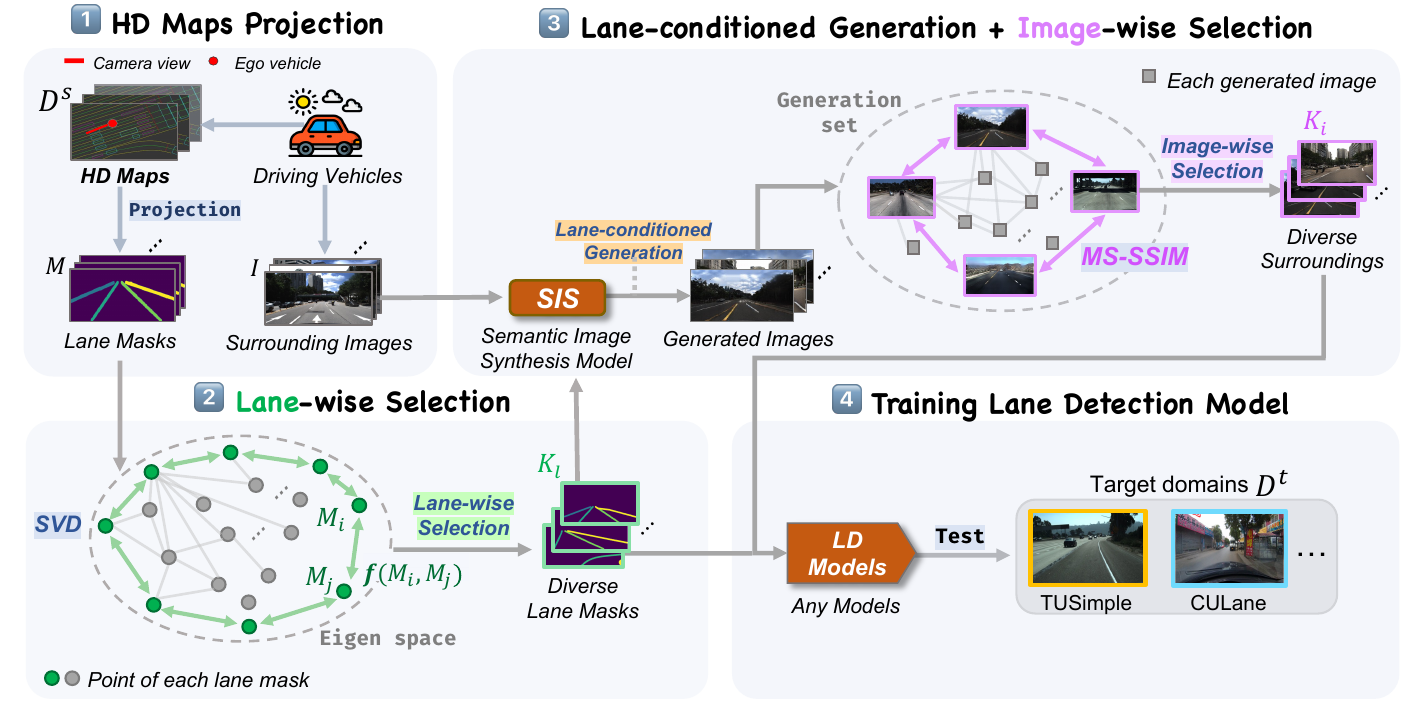}
    \caption{\textbf{The overall framework.} Our framework adopts HD Maps as a single-source domain to build robust lane detection models. 
    We first extract numerous lane masks from HD Maps with surrounding images. Then, we train SIS (Semantic Image Synthesis) generative models to synthesize lane-conditioned images with diverse surroundings. 
    We select a core set among synthesized images with two criteria to maximize its diversity: \textit{(i) lane structure} and \textit{(ii) surroundings}. 
    Then, we finally train the lane detection model and obtain enhanced generalization performance.
    }
    \label{fig:fig2_overall_arch}
    \vspace{-5mm}
\end{figure*}

\textbf{Problem Statement.}
For the Single-Source Domain Generalization (SSDG) in lane detection, we use source domain ${\bm{D^s}=\left\{(I_i, M_i)\right\}_{i=1}^N}$ which consists $N$ pairs of front-view (i.e., \textit{surrounding}) image $I$ and lane mask $M$. 
Notably, our work demonstrates HD Maps provide a successful source for mitigating domain discrepancy in lane detection. 
To train the lane detection model as robustly as possible to the unseen target domains ${\bm{D^t}}$, we propose a diversity-aware framework by selecting a core set for the lane and surrounding. 

\subsection{Diversity-aware Coreset Selection Criteria}\label{sec:subsec:core}

We first introduce a coreset selection method to boost the diversity of a single domain. 

\textbf{Data Selection Policy.}  
In this work, we define \textit{maximal diversity} as \textit{minimum sum of similarities} between nodes in a graph.
Let's assume a complete weighted graph $G=(V, E, W)$ where the vertex set $V=\{v_1, v_2, ..., v_N\}$ represents $N$ data points, and each data point is connected to every other data point through an edge $E=\{(v_i, v_j)\}$. The weight of each edge $W=\{w(i,j)\}$ is defined as the similarity between the two data points connected by the edge. The near-optimal solution can be configured as Algorithm~\ref{alg:search_algo}.   
This is freely applied to various types of data by defining each type of metric, hence it is easily expanded to both lane structure and surrounding image.   
Therefore, we define two similarity criteria for \textit{(i) lane} and \textit{(ii) surrounding images} as follows.

\textbf{\textit{(i) Criteria for Lane Structure}.} 
We propose a lane similarity function $f$ between two lane masks $\textbf{M}_i, \textbf{M}_j$ that consists of several lanes $\ell$ with its curvatures. 
Inspired by Eigenlanes~\cite{eigenlane}, we project each lane $\ell$ onto the image plane and compare the similarity represented in pixel coordinates. 
More specifically, we express a lane as a set of 2D points uniformly sampled along the vertical direction in the image plane, i.e., $\ell=[x_1, x_2, ..., x_p]^T$, where $x_i$ denotes the $x$-coordinate of the $i$-th sample, and $P$ is the number of points. 
Then, we construct a lane matrix $\bm{A}=[\ell_1, \ell_2, ..., \ell_L]$ by utilizing all $L$ lane curves from the entire training data. 
By applying the SVD to the constructed lane matrix $\bm{A}\in\mathbb{R}^{P\times L}$, we factorize it as follows:
\begin{equation}
   \bm{A}= \bm{U} \Sigma \bm{V}^T,
\end{equation}
where $\bm{U}=[\textbf{u}_1, ..., \textbf{u}_P] \in\mathbb{R}^{P\times s}$ and $\textbf{V}=[\textbf{v}_1, ..., \textbf{v}_L] \in\mathbb{R}^{s\times L}$ are orthogonal matrices, and $\Sigma\in\mathbb{R}^{s\times s}$ is a singular value matrix. 

Using this, each lane $\ell$ is approximated as $\tilde{\ell}$ using $R$-rank approximation with singular vectors $\textbf{u}_1, ..., \textbf{u}_R$ as follows:
\begin{equation}
\tilde{\ell_i} = \textbf{U}_R \textbf{c}_i = [\textbf{u}_1, ..., \textbf{u}_R] \textbf{c}_i.
\end{equation}
where $\textbf{U}_R \in\mathbb{R}^{P\times R}$ . 
The coefficient vector $\textbf{c}_i$ of each lane $\ell_i$ in the \textit{eigenlane} space spanned by the singular vectors $[\textbf{u}_1, ..., \textbf{u}_R]$ is expressed as follows:
\begin{equation}
\textbf{c}_i = \textbf{U}_{R}^{T} \ell_i = [c_1, ..., c_R].
\end{equation}

Therefore, we simply define the similarity between two lanes $\ell_i$ and $\ell_j$ as Euclidean distance $d$ by projecting them onto the \textit{eigenlane} space as follows:
\begin{equation}
d(\ell_i, \ell_j) = | \textbf{c}_i - \textbf{c}_j |.
\end{equation}

Moreover, our goal is to extend this lane-wise similarity to lane masks.
Given two lane masks $\textbf{M}_i=[\ell_{0}^{i}, ..., \ell_{m}^{i}]$ and $\textbf{M}_j=[\ell_{0}^{j}, ..., \ell_{n}^{j}]$, each containing $m$ and $n$ lanes in their lane masks, respectively.
Assuming that $m\ge n$ always holds, we define the similarity function $f$ between lane masks $\textbf{M}_i, \textbf{M}_j$ as follows: 
\begin{equation}
  f(\textbf{M}_i, \textbf{M}_j) = \sum^{n}_{k} \min  d( \ell_{l}^{i}, \ell_{k}^{j} )  + |m-n|~\kappa,
\end{equation}
where $\kappa$ denotes the penalty constant which we describe next. 
We believe that the number of lanes is also an important factor in determining the similarity. 
Therefore, when two-lane masks contain different numbers of lanes, we penalize their similarity by adding hyperparameter $\kappa$. 
With the above definition of similarity between lane masks, we select the $K_{\ell}$ number of lane masks by using the simple selection algorithm in Algorithm~\ref{alg:search_algo}.

\begin{algorithm}[t]
\small
\caption{Data Selection Policy}
\begin{algorithmic}[1]
\Require Complete weighted graph $G$ and $K=\#$ of vertices
\State Find the vertex pair $(i,j)$ which has minimum $w(i, j)$
\State Add $i$ and $j$ to $Q$
\While{$|Q| < K$}
\State Find vertex $k$ which minimizes $\sum_{l \in V \setminus Q} w(k, l)$
\EndWhile
\Ensure $Q$
\end{algorithmic}
\label{alg:search_algo}
\end{algorithm}



\textbf{\textit{(ii) Criteria for Surrounding Image}.}
By training the model on RGB images that have the \textit{same} lane masks but \textit{different} image surroundings, we aim to make the model less sensitive to variations in those surroundings.
However, unlike the case of lane layouts, it is difficult to obtain a quantitative similarity metric between surroundings of RGB images. 
Therefore, we adopt a comparison metric commonly used to check the quality of generated images by image synthesis models.

The Structural Similarity Index Measure (SSIM)~\cite{ssim} and its Multi-Scale extension (MS-SSIM)~\cite{ms-ssim} are common metrics to evaluate the perceived quality or similarity of images.  
Since MS-SSIM has also been used to compare the quality of generated images against the reference images~\cite{snell2017learning}, we choose the MS-SSIM as the similarity metric between two surrounding images.     
Therefore, we select $K_{i}$ surrounding images using MS-SSIM using the Algorithm~\ref{alg:search_algo}.

\subsection{Overall Framework}\label{sec:subsec:train}
Combining with our coreset selection criteria, we now introduce a diversity-aware framework to raise the robustness of the lane detection model. 

\textbf{Extracting Lane Masks with Lane Selection.} We note that the huge potential of HD Maps is the flexibility of extracting diverse lane masks according to camera parameters. 
Using them to pre-built HD Maps, we extract the pair of lane masks $M$ with their surrounding image $I$ by projecting the lane objects onto the camera image plane at the current location. 
It is worth noting that this process does not require any additional annotation to obtain the pixel-wise labels. 
Also, with this projection, we acquire a substantial number of lane masks from HD Maps easily.

After getting these $N$ pairs of lane masks and surrounding images, we adopt the aforementioned lane-wise selection criteria to boost lane diversity.  Finally, we select $K_{\ell}$ number of lane masks among $N$ where  $K_{\ell}<N$. 

\textbf{Lane-conditioned Generation with Image Selection.}
Next, utilizing these $K_{\ell}$ selected lane masks, we adopt Semantic Image Synthesis (SIS) methods~\cite{sushko2020you, wang2018high} to generate lane-conditioned images. 
Unlike conventional SIS approaches, we use only lane masks as a condition to generate images to make diverse surroundings. 
We generate 100 RGB images for each lane label mask. 

After that, we now adopt the aforementioned image selection criteria to select diverse surroundings among generated images. We finally select $K_{i}$ number of generations. 

\textbf{Training Lane Detection Model with Coreset.}
Combining selections from both lane masks and surroundings, we build the coreset which maximized diversity. Please note that this coreset is originally from HD Maps. 
Then, we finally train the lane detection model using this dataset. 
It is noteworthy that our framework is model-agnostic, which means it applies all lane detection models like data augmentation.  



\section{Experiments}

\begin{table*}
\vspace{0.3cm}
\centering
\renewcommand{\arraystretch}{1.1}
\setlength{\tabcolsep}{.45em}    
    \caption{\textbf{Quantitative results on CULane target domain}.
    \dag Note that we attempted to reproduce \cite{li2022multi}, but were not even able to reproduce their performance. In this table, we present their results as they are reported.
    }
    \resizebox{\textwidth}{!}{%
    \begin{tabular}{l|ccc|cccccccccc}
    \hline
    \multicolumn{1}{c|}{\multirow{2}{*}{Experiment setting}} & \multicolumn{3}{|c|}{Trained data} & \multirow{2}{*}{Total} & \multirow{2}{*}{Normal} & \multirow{2}{*}{Crowded} & \multirow{2}{*}{Night} & \multirow{2}{*}{No line} & \multirow{2}{*}{Shadow} & \multirow{2}{*}{Arrow} & \multirow{2}{*}{Dazzle light} & \multirow{2}{*}{Curve} & \multirow{2}{*}{Crossroad} \\ \cline{2-4}
    
    \multicolumn{1}{c|}{} & TUSimple & CULane & LabsLane &  &  &  &  &  &  &  &  &  &  \\ \hline
    
    Source only - ERFNet & \ding{52} &  &  & 24.2 & 41 & 19.6 & 9.1 & 12.7 & 12.7 & 28.8 & 11.4 & 32.6 & 1240 \\
    Source only - GANet & \ding{52} &  &  & 30.5 & 50 & 25.5 & 19.6 & 18.5 & 16.8 & 36 & 25.4 & 34.2 & 7405 \\ 
    Advent~\cite{advent} & \ding{52} & \ding{52} &  & 30.4 & 49.3 & 24.7 & 20.5 & 18.4 & 16.4 & 34.4 & 26.1 & 34.9 & 6527 \\
    PyCDA~\cite{pycda} & \ding{52} & \ding{52} &  & 25.1 & 41.8 & 19.9 & 13.6 & 15.1 & 13.7 & 27.8 & 18.2 & 29.6 & 4422 \\
    Maximum Squares~\cite{maximum} & \ding{52} & \ding{52} &  & 31 & 50.5 & 27.2 & 20.8 & 19 & 20.4 & 40.1 & 27.4 & 38.8 & 10324 \\
    MLDA~\cite{li2022multi}\dag & \ding{52} & \ding{52} &  & \underline{38.4} & \textbf{61.4} & \textbf{36.3} & \underline{27.4} & \underline{21.3} & \textbf{23.4} & 49.1 & \underline{30.3} & \underline{43.4} & 11386 \\ 
   {\cellcolor[HTML]{E6F4FA}Source only - GANet} & \cellcolor[HTML]{E6F4FA} & \cellcolor[HTML]{E6F4FA} & \cellcolor[HTML]{E6F4FA}\ding{52} & \cellcolor[HTML]{E6F4FA}34.9 & \cellcolor[HTML]{E6F4FA}52.4 & \cellcolor[HTML]{E6F4FA}25.9 & \cellcolor[HTML]{E6F4FA}27.1 & \cellcolor[HTML]{E6F4FA}18.8 & \cellcolor[HTML]{E6F4FA}20.3 & \cellcolor[HTML]{E6F4FA}44.5 & \cellcolor[HTML]{E6F4FA}25.9 & \cellcolor[HTML]{E6F4FA}42.7 & \cellcolor[HTML]{E6F4FA}\textbf{939} \\  
    {\cellcolor[HTML]{E6F4FA}Ours - ERFNet} & \cellcolor[HTML]{E6F4FA} & \cellcolor[HTML]{E6F4FA} & \cellcolor[HTML]{E6F4FA}\ding{52} & \cellcolor[HTML]{E6F4FA}38.2 & \cellcolor[HTML]{E6F4FA}\underline{56.8} & \cellcolor[HTML]{E6F4FA}32.5 & \cellcolor[HTML]{E6F4FA}24.4 & \cellcolor[HTML]{E6F4FA}\textbf{23.8} & \cellcolor[HTML]{E6F4FA}\underline{22.3} & \cellcolor[HTML]{E6F4FA}\underline{51.2} & \cellcolor[HTML]{E6F4FA}\textbf{32.2} & \cellcolor[HTML]{E6F4FA}40.9 & \cellcolor[HTML]{E6F4FA}4572 \\
    {\cellcolor[HTML]{E6F4FA}Ours - GANet} & \cellcolor[HTML]{E6F4FA} & \cellcolor[HTML]{E6F4FA} & \cellcolor[HTML]{E6F4FA}\ding{52} & \cellcolor[HTML]{E6F4FA}\textbf{39.6} & \cellcolor[HTML]{E6F4FA}56.7 & \cellcolor[HTML]{E6F4FA}\underline{34} & \cellcolor[HTML]{E6F4FA}\textbf{30.1} & \cellcolor[HTML]{E6F4FA}21.2 & \cellcolor[HTML]{E6F4FA}22 & \cellcolor[HTML]{E6F4FA}\textbf{53.2} & \cellcolor[HTML]{E6F4FA}27.1 & \cellcolor[HTML]{E6F4FA}\textbf{46.9} & \cellcolor[HTML]{E6F4FA}\underline{1275} \\  \hline
    \end{tabular}%
    }
    \label{ref:table_qualitative_result}
\end{table*}

\subsection{Datasets}
\textbf{LabsLane (HD Maps).} We use Naver Labs HD Maps\footnote{https://naverlabs.com/en/datasets} to create lane label masks, which are publicly available for research purposes and provide rich information about the road environment in cities of South Korea.
During the projection of the lane object onto the camera view, we only consider a maximum of four lanes near the ego vehicle, which is consistent with other benchmark datasets that also have a maximum of four or five lanes.
As a result, the generated dataset, dubbed LabsLane, consists of 7,184 RGB images and corresponding 6 DoF poses.

\textbf{CULane.} It consists of complex \textit{urban road environments} and 133,235 frames, of which 88,880, 9,675, and 34,680 are used for training, validation, and testing, respectively. 

\textbf{TUSimple.} It mostly consists of \textit{highway} driving scenes and around 7,000 video clips, each containing 20 frames, with labeled lanes on the final frame of each clip.

\subsection{Implementation details}
We used GANet-small~\cite{wang2022keypoint} and ERFNet~\cite{romera2017erfnet} as lane detection models which are respectively keypoint-based and segmentation-based methods. 
For the ERFNet setting, we faithfully followed the MLDA~\cite{li2022multi} setup.
We use the F1 score and accuracy respectively to measure the performance metric for CULane and TUSimple like the official evaluation code. 

\begin{table}[t!]
\centering
\caption{\textbf{Quantitative results on TUSimple target domain.} 
Our framework using LabsLane shows similar performance with Target only.}
\label{tab:ablation_study}
\renewcommand{\arraystretch}{1.1}
\resizebox{\columnwidth}{!}{%
    \begin{tabular}{l|ccc|ccc}
    \hline
    \multicolumn{1}{c|}{} & \multicolumn{3}{c|}{Trained data} &  &  &  \\ \cline{2-4}
    \multicolumn{1}{c|}{\multirow{-2}{*}{Experiment setting}} & CULane & TUSimple & LabsLane & \multirow{-2}{*}{Acc} & \multirow{-2}{*}{FP} & \multirow{-2}{*}{FN} \\ \hline
    Source only - ERFNet & \ding{52} &  &  & 60.9 & 31.6 & 55.2 \\
    Source only - GANet & \ding{52} &  &  & 76.9 & {\ul 25.8} & 35 \\ 
    Target only - GANet &  & \ding{52} &  & 95.9 & 1.97 & 2.62 \\ \hline
    Advent~\cite{advent} & \ding{52} & \ding{52} &  & 77.1 & 39.7 & 43.9 \\
    PyCDA~\cite{pycda} & \ding{52} & \ding{52} &  & 80.9 & 51.9 & 45.1 \\
    Maximum Squares~\cite{maximum} & \ding{52} & \ding{52} &  & 76 & 38.2 & 42.8 \\
    MLDA~\cite{li2022multi} & \ding{52} & \ding{52} &  & {\ul 89.7} & 29.5 & {\ul 18.4} \\ 
    \rowcolor[HTML]{E6F4FA} 
    \rowcolor[HTML]{E6F4FA} 
    Ours  &  &  & \ding{52} & \textbf{92.4} & \textbf{6.7} & \textbf{9.4} \\ \hline
    \end{tabular} %
    }
    \label{ref:table_tusimple}
    \vspace{-0.6cm} 
\end{table}

\subsection{Main results}
\textbf{Results on CULane.}    
In \Cref{ref:table_qualitative_result}, we summarize the performances of each lane detection model on CULane dataset, which includes various scenario cases. 
We compared our methods with domain adaptation methods such as Advent~\cite{advent}, PyCDA~\cite{pycda}, Maximum Squares~\cite{maximum}, and MLDA~\cite{li2022multi}. 
Although our approach is an SSDG setup, we observe that our total performance outperforms MLDA.
In particular, the performance improvement is greater when it is not a typical surrounding scene (e.g. Night, Dazzle light). 
This demonstrates that our HD Map-based diversity maximization framework robustly handles large domain shifts.

\textbf{Results on TUSimple.}
We also validate our proposed framework on the TUSimple dataset in \Cref{ref:table_tusimple}.
We found that our framework also outperforms the DA-based methods even though we do not use target domain information.
It is especially notable that both MLDA~\cite{li2022multi} and the proposed method showed significant performance improvements compared to the results on the CULane dataset.

\begin{figure}[t!]
    \vspace{0.2cm}
    \centering
    \includegraphics[width=0.43\textwidth]{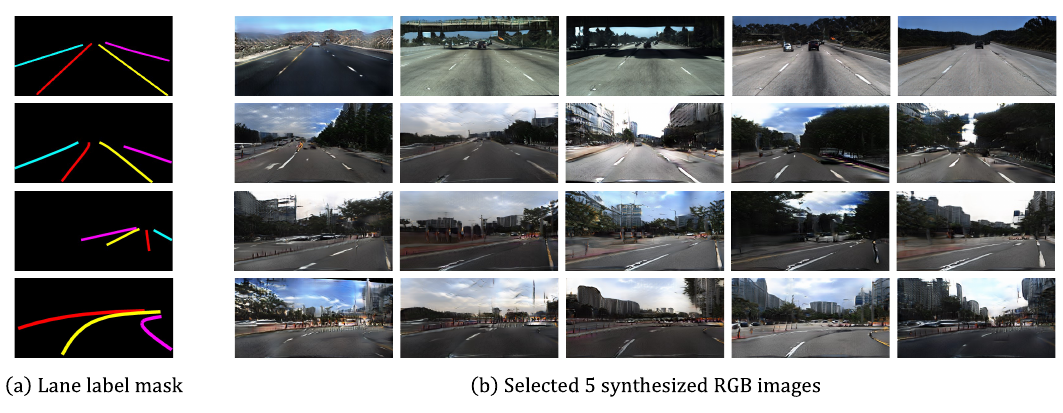}
    \caption{\textbf{Examples of generated dataset.} We utilized OASIS~\cite{sushko2020you} as our generative model to visualize results. 
    }
    \label{fig:fig3_examples}
\end{figure}

\textbf{Generated Images from SIS.}
In \Cref{fig:fig3_examples}, we show the generated images from simple SIS models OASIS~\cite{sushko2020you}. Our framework succeeds in generating diverse surroundings from numerous lane markings from HD Maps. 
We believe there is more room for improvement when we adopt diffusion-based models.

\section{Conclusion}
In this paper, we propose a novel training framework that utilizes HD Maps for robust lane detection. We generated a large number of images from HD Maps with varying surroundings and lane structures to mitigate domain discrepancy on lane detection. 

{
    \small
    \bibliographystyle{ieeenat_fullname}
    \bibliography{main}
}


\clearpage
\setcounter{page}{1}
\maketitlesupplementary

\section{Domain discrepancy in Lane Detection}

\begin{figure}[t!]
    \vspace{0.2cm}
    \centering
    \includegraphics[width=0.5\textwidth]{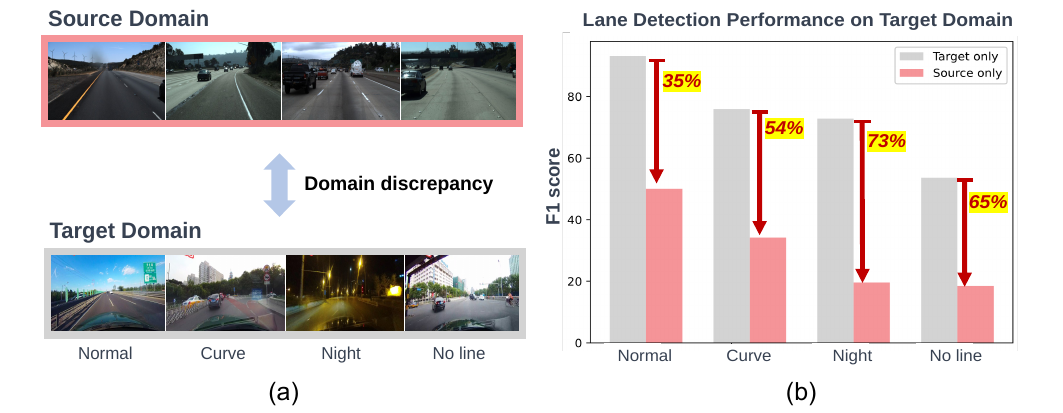}
    \caption{\textbf{Performance degradation by domain discrepancy.} 
    (a) Example images of source and target domains and
    (b) performance of the state-of-the-art lane detection model~\cite{wang2022keypoint} on the target domain, still revealing significant challenges due to domain differences. 
    }
    \label{fig:intro}
\end{figure}

We explored the cause of domain discrepancy that occurs in lane detection environments. It recurs to severe performance degradation illustrated in \Cref{fig:intro}. After we observed various factors, we divided the cause of the above domain discrepancy into \textit{(i) lane structures} and \textit{(i) surroundings}. 

\begin{figure}[!]
    \centering
    \subfloat[]{\includegraphics[width=0.645\linewidth]{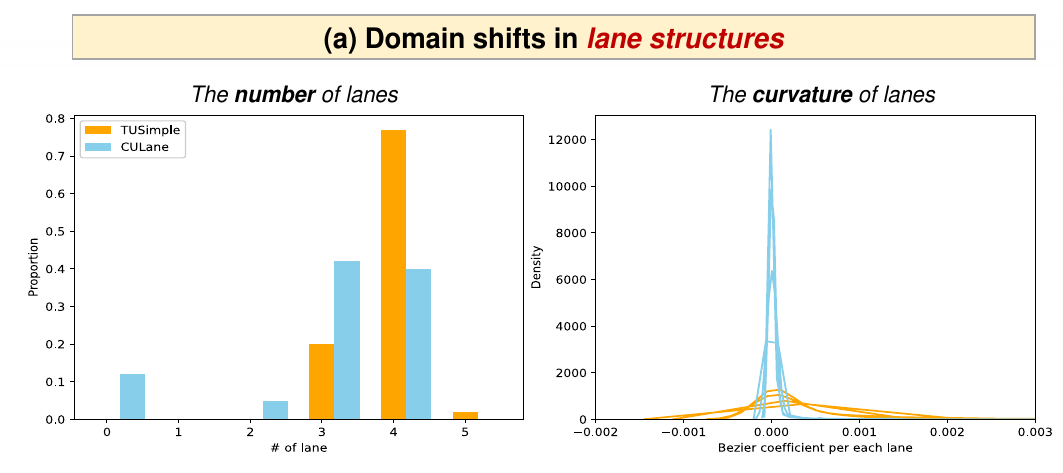}}
    \subfloat[]{\includegraphics[width=0.355\linewidth]{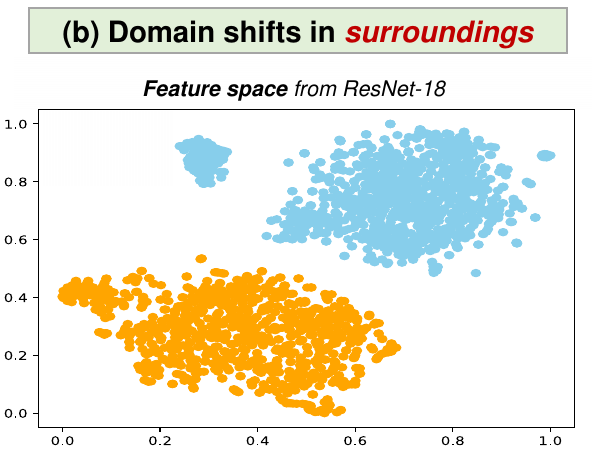}}
    \caption{\textbf{Domain discrepancy in lane detection.} We categorized the cause of domain discrepancy in lane detection. (a) The difference in the number of lanes and curvatures shows a clear discrepancy in lane structures. (b) Judging from the t-SNE~\cite{tsne} analysis, the contextual difference of surroundings is also confirmed.}
    \label{fig:domain_diff}
    \vspace{-0.4cm} 
\end{figure}

For lane structures, we compared the number of lanes included in each scene and curvatures as shown in Fig~\ref{fig:domain_diff}-(a). 
Especially, for detecting lane curvatures, we used the pre-released bezier control points~\cite{feng2022rethinking} to calculate the bezier coefficients which represent the curvature of lanes. 
These plots faithfully demonstrate that the difference in lanes can be one of the factors of the domain discrepancy in lane detection scenarios. 
In addition, as shown in Fig~\ref{fig:domain_diff}-(b), we conducted a t-SNE~\cite{tsne} analysis to observe differences in surroundings. We first extracted image features from ResNet-18 and reduced the dimension via t-SNE. As shown in the result, images from the two datasets form distinctly separate clusters in the feature space. 

\section{More Ablations}

\textbf{Using benchmark images instead of HD maps:}    
We also conducted experiments on benchmark images rather than HD maps to verify the efficacy of our training framework itself. 
We measured the performance of the model learned from TUSimple to CULane.
Note that it is different from MLDA~\cite{li2022multi} setting which targets Domain Adaptation scenario. 
As shown in Table~\ref{ref:tu2cu}, our framework outperforms the source-only model, especially when our selection method is employed. Also, note that the performance increased for both Pix2pix and OASIS. We expect any semantic image synthesis model can be easily utilized in our framework


\begin{figure}[!]
    \vspace{0.2cm}
    \centering
    \includegraphics[width=0.28\textwidth]{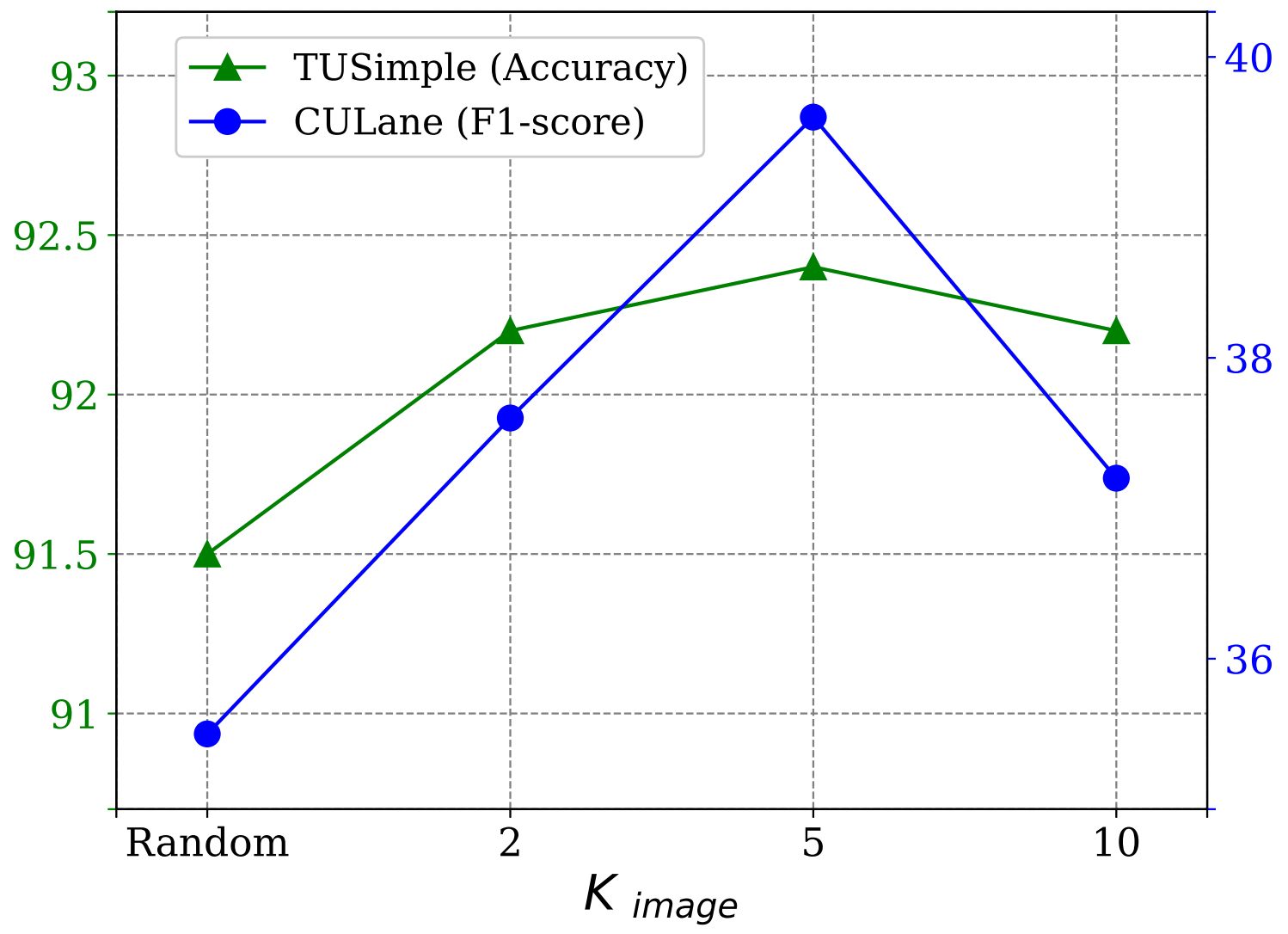}
    \caption{
    \textbf{Effect of the number of selected data}.
    We compare the performance of randomly selecting 10 data from 100 images generated from a single lane label mask and selecting 2, 5, and 10 data using the proposed diversity-aware image selection method.
    }
    \label{fig:efficacy_selection}
\end{figure}

\textbf{Effect of the number of selected images:}
We also experimented with the efficacy of the number of selected images.
We used accuracy and total F1-score as comparison metrics for TUSimple and CULane datasets, respectively.
In Fig.~\ref{fig:efficacy_selection}, we compare the performance of randomly selecting 10 samples from a single lane label mask of LabsLane dataset with selecting 2, 5, and 10 samples using the proposed diversity-aware image selection method.
The figure shows that the proposed selection method generally outperforms random selection, and the performance is the highest when selecting 5 data for each lane label.
Therefore, we set the number of synthesized data $K_{image}$ to 5 for all experiments.

\begin{table}[t!]
\centering
\renewcommand{\arraystretch}{1.3}
\begin{adjustbox}{width=0.4\textwidth}
\resizebox{\columnwidth}{!}{%
    \begin{tabular}{lccc}
    \hline
    \multicolumn{1}{c}{} & Total & Normal & Curve \\ \hline
    Source only (TUSimple) & 24.1 & 41 & 32.6 \\ \hline
    Pix2pix~\cite{wang2018high} & 26.7 & 43 & 33.2 \\
    OASIS~\cite{sushko2020you} & 27.3 & 45 & 37.3 \\
    \rowcolor[HTML]{EFEFEF} 
    OASIS + Our selection method & \textbf{31.2} & \textbf{47.9} & \textbf{40.4} \\ \hline
    \end{tabular}%
    }
    \end{adjustbox}
\caption{\textbf{Evaluating our methods using TUSimple images.} To verify the versatility of our framework, we tested our framework on benchmark dataset.}
\label{ref:tu2cu}
\vspace{-.4cm}
\end{table}

\textbf{Using other image synthesis model:}
We also validated that our proposed framework works well with other image synthesis models.
We used Pix2PixHD~\cite{wang2018high}, another algorithm that can generate diverse images by re-sampling noise vectors. 
We trained the image synthesis model on TUSimple and evaluated the performance on total F1-score of CULane. 
As shown in Table~\ref{ref:tu2cu}, we observed a performance improvement of about 10\% when using Pix2PixHD~\cite{wang2018high} compared to training on TUSimple.
However, we believe that the reason for the difference in performance is due to the quality of the generated images, as OASIS~\cite{sushko2020you} produced more realistic images compared to Pix2PixHD~\cite{wang2018high}.

\section{Discussions}

\begin{table}[t!]    
    \centering
    \renewcommand{\arraystretch}{1.3}
    \begin{adjustbox}{width=0.3\textwidth}
    \resizebox{\columnwidth}{!}{%
        \begin{tabular}{l|cc|c}
        \hline
        \multicolumn{1}{c|}{} & \multicolumn{2}{c|}{Source domain} &  \\
        \multicolumn{1}{c|}{\multirow{-2}{*}{}} & Simulator & HD map & \multirow{-2}{*}{Acc} \\ \hline
        UNIT~\cite{simunit} & \ding{52} &  & 77.5 \\
        MUNIT~\cite{simmunit} & \ding{52} &  & 78.6 \\
        ADA~\cite{simada} & \ding{52} &  & 82.9 \\ \hline
        \rowcolor[HTML]{EFEFEF} 
        Ours &  & \ding{52} & \textbf{92.4} \\ \hline
        \end{tabular}%
    }
    \end{adjustbox}
\caption{\textbf{Results on TUSimple compared to Sim-to-Real domain adaptation.} Our performance outperforms the existing Sim-to-Real DA performance without taking advantage of any information from TUSimple.}
\label{ref:sim2ours}
\vspace{-.4cm}
\end{table}

\textbf{Comparison with Sim-to-Real adaptation method:} 
The framework using HD map can be regarded as similar to the simulator in that it generates diverse augmented data. Therefore, we compared the performance on TUSimple with existing Sim-to-Real domain adaptation models~\cite{simunit, simmunit, simada} in Table~\ref{ref:sim2ours}. 
Their photo-realistic source domain was made by CARLA~\cite{simcarla} which is a gaming engine to generate synthetic datasets. 
Note that, UNIT~\cite{simunit}, MUNIT~\cite{simmunit}, and ADA~\cite{simada} models included TUSimple's unlabeled image as they are domain adaptation works. However, our result outperforms the existing Sim-to-Real DA performance without taking advantage of any information from TUSimple.

\textbf{Comparison with cross-domain adaptation method:} 
In Table \ref{ref:table_qualitative_result}, we compared the proposed framework with domain adaptation methods. 
We confirmed that our proposed framework shows comparable results with MLDA even without any knowledge of the target domain. 
Furthermore, we believe that the proposed framework is not limited to a specific lane detection model but can be applied to other various lane detection models as well.


\end{document}